\documentclass[runningheads]{llncs}
\usepackage[T1]{fontenc}
\usepackage{times}
\usepackage{latexsym}

\usepackage{amsmath,graphicx}
\usepackage{multirow}
\usepackage{hyperref}
\usepackage{booktabs}

\newcommand{\modelname}{SF-Re2G}

\usepackage{graphicx}
\begin{document}
\title{Structures Facilitate Retrieve, Rerank, and Generate}
%
%
\author{Anonymous Author(s)}  
\authorrunning{Anonymous Author(s)}

\institute{
Anonymous Institution(s) \\
\email{anonymous@example.com}  
}

\author{Yeqin Zhang\inst{1,2} \and
Haomin Fu\inst{1,2}\and
Xujie Zhang\inst{1,2} \and
Cam-Tu Nguyen \inst{1,2*}}
\institute{State Key Laboratory for Novel Software Technology, Nanjing University, China \and
School of Artificial Intelligence, Nanjing University, China
\email{zhangyeqin@smail.nju.edu.cn}}

\maketitle              
\begin{abstract}
Document-grounded dialogue systems (DGDS) utilize knowledge from external documents to answer domain-specific user questions. Existing solutions typically divide documents into independent passages for retrieval and response generation. This approach, however, neither makes good use of structural information within documents nor provides enough (document) context for knowledge selection and responses. This paper proposes {\modelname} to address such issues systematically. Firstly, we seek to improve a passage representation by contrasting it with others of the same section, thus improving the retrieval performance. Secondly, a structure-enhanced reranker is built, leveraging the fact that multiple grounding passages of one dialog turn tend to be in the same neighborhood. Specifically, candidates from the retrieval are grouped into subgraphs according to the document structure. The reranker will rescore the candidate integrating its group information. Finally, the chosen passages are used for responses, taking into account the subgraph context for better generation. Experimental results on two DGDS datasets validate our method for both Chinese and English. 

\keywords{Structure-Aware Retrieval  \and Contrastive Learning \and Passage Reranking.}
\end{abstract}
\section{Introduction}

\begin{figure}
    \centering
    \includegraphics[width=0.98\textwidth]{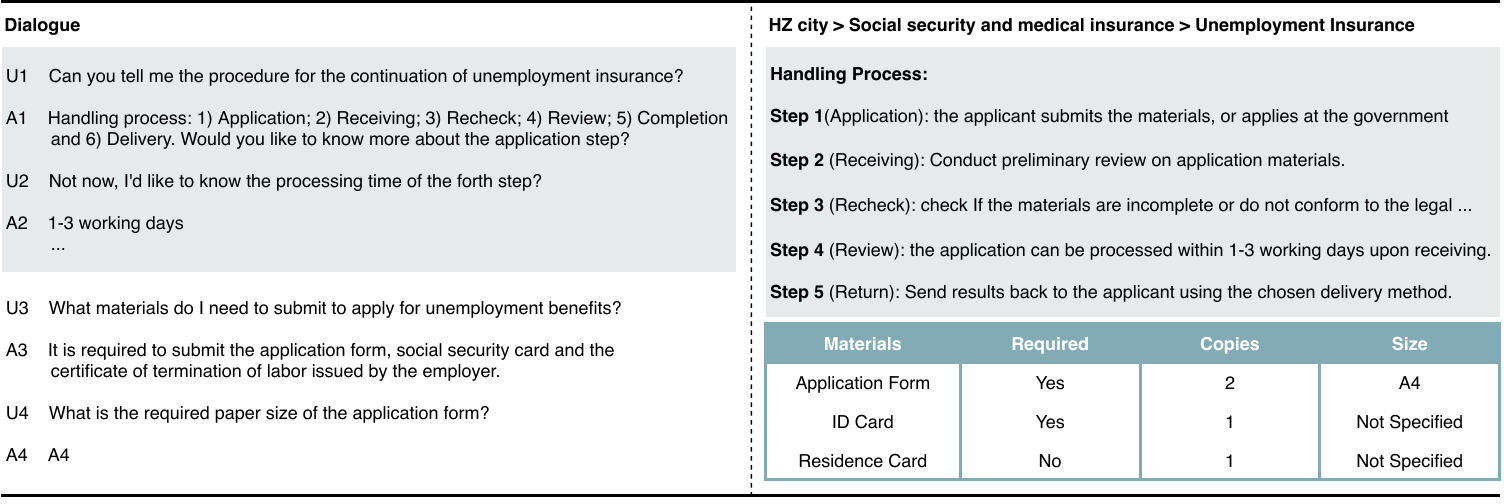}
    \caption{An example from Doc2Bot dataset. The left is a dialogue, and the right is a grounding document. The document is divided into several passages. Each step in the document and each cell in the table is treated as a passage.}
    \label{fig:example}
\end{figure}

Document-grounded dialogue systems (DGDS) \cite{feng2020doc2dial,ma2021unstructured,feng2021multidoc2dial} is a hot topic in natural language processing (NLP) research, in which a set of external documents the agent should base on to generate the response. By doing this, more meaningful and accurate responses can be generated by the dialog system. DGDS differs from traditional QA tasks in three aspects: multi-turn, close-domain, and document structure. Multi-turn results in the fact that context is longer, and the interaction between context and candidate response must be considered. Close-domain makes the passages in the documents semantically related. This requires the knowledge selection module to focus on similar passages for response generation. Unlike the structured knowledge base and the knowledge graph, document structure is usually a tree structure formed by organizing the document and the knowledge in the document according to certain semantics or logic. The more semantically similar the two knowledge, the closer they are in the hyperbolic space \cite{DBLP:conf/chi/LampingRP95} 
 formed by this document structure tree. When multiple pieces of knowledge are needed to answer the user's query, the distance between these pieces of knowledge is usually small in this space.

Existing solutions \cite{fu-2022-towards,feng-etal-2022-multi} exploit the commonly used methods for open QA, ignoring the distinct characteristics of DGDS. Specifically, RAG \cite{patrick2020rag} and Re2G \cite{glass-etal-2022-re2g} are utilized for knowledge retrieval and dialog generation. These methods, however, treat documents as textual sequences, which are then divided into independent passages for knowledge retrieval and response generation. Consequently, this gives raise to two main issues when applying to DGDS. First, it is nontrivial to have the right choice of granularity for passages. Long passages may contain more noise, whereas finer passages do not provide enough context for matching and generation. Second, the inherent structures within the grounding documents are ignored, which might result in suboptimal retrieval and generation. 

Figure \ref{fig:example} illustrates the aforementioned issues with an example dialog from Doc2Bot \cite{fu2022doc2bot} dataset. First, it is not easy to have the right choice of passage length. For question \texttt{U1}, the system needs to have an overview of the whole handling process, which requires the knowledge from a large span of texts. On the other hand, for question \texttt{U2}, we can just focus on the finer knowledge of Step 3 on the document. Having a long context may produce too much noise for the question \texttt{U2}, but is better for question  \texttt{U1}. Second, the structure information is essential for some complex structure. As you can see from Figure \ref{fig:example}, questions from \texttt{U3} to \texttt{U4} are grounded on knowledge from a table. Although we can sequentialize the whole table for dense passage retrieval as in RAG and Re2G, we then lose the semantic connections between such as ``Materials''-``Application Form''.

 To address these issues, we do not fix a passage length but exploit nodes in the document graph as passages retrieval. Note that, doing so results in passages of different length. For example, those that are in tables might be very short, and those corresponding to paragraphs are relatively longer. We then try to better integrate the context by exploiting document structures. 
 
 Specifically, we propose a novel approach \modelname{} that leverages structural information to facilitate the full DGDS process, including passage retrieval, reranking, and response generation. To the best of our knowledge, this is the first work to systematically study the impact of structure on the commonly used DGDS pipeline. In terms of retrieval, we make two improvements to the deep neural passage retrieval model DPR \cite{karpukhin2020dpr}. Inspired by TreeJC \cite{wan2021treejc}, we design a top-down aggregation to enhance the representation of passages using upper-level information such as paragraph headings and document names. Additionally, we recognize that passages under the same section in the document graph tend to have similar content, which makes them more difficult to be distinguished by the model. Therefore, we introduce a structural contrastive approach, which encourages the model to focus more on neighboring passages. For reranking, we dynamically construct some groups based on retrieved passages according to the document structure. Then we model the probability of candidate passages given the user's query and the same group passages, producing a structurally enhanced reranking score. And during the generation process, we integrate this enhanced score into the generator using a weighting method similar to RAG \cite{patrick2020rag} and Re2G \cite{glass-etal-2022-re2g}.

We conduct experiments on the MultiDoc2Dial dataset \cite{feng2021multidoc2dial} and Doc2Bot dataset \cite{fu2022doc2bot}, which have been recently introduced to build document grounded dialogue systems. We re-construct the document graphs for each knowledge document based on the original data. The experimental results prove the superiority of our method. The main contributions of this work are: 
\begin{itemize}
    \item We reveal that structural information is very valuable and not negligible for DGDS. Current methods that ignore these structural elements can result in suboptimal performance.
    \item We introduce a novel method \modelname{}, which integrates the structural information for all stages of DGDS. 
    \item Our method has a superior performance compared to other state-of-the-art approaches in recent years.
\end{itemize}

\section{Related Works}
The primary DGDS pipeline can be simply divided into knowledge selection and response generation. Existing studies optimize one of these modules individually or optimize all of them as a whole.

\subsection{Knowledge Selection}
Extensive studies have proposed methods to improve knowledge selection from different aspects. BM25 \cite{robertson2009bm25} uses traditional match-based methods to retrieve passages. DPR \cite{karpukhin2020dpr} is an early attempt for dense passage retrieval by deep neural networks.  For a detailed discussion of other knowledge selection methods, please refer to S1.1 of the supplementary materials.

\subsection{Multi-turn Dialog Representation}
Some studies focus on multi-turn dialog matching. DAM \cite{zhou2018multi} matches a candidate response to different granularities of the context (words, phrases, sentences). \cite{tao-etal-2019-one} performs deep interactions between the context and response by stacking multiple interaction blocks. Additional multi-turn dialog representation approaches are presented in S1.2 of the supplementary materials, along with a discussion on the inefficiency of retrieval models.

\subsection{Knowledge-enhanced Generation}
There are several mainstream approaches combining knowledge to generation. Fusion-in-Decoder \cite{izacard-grave-2021-leveraging} technique concatenates the encoder outputs of retrieved passages as input for the decoder. RAG \cite{patrick2020rag} and Re2G \cite{glass-etal-2022-re2g} take passage relevance scores to weight generation outputs. UniGDD \cite{gao-etal-2022-unigdd} concatenates the texts of retrieved passages as input for the encoder. \cite{li-etal-2019-incremental} proposed 2-stage decoders: the first generates output based on the context, and the second refines the response by integrating the document. \cite{zhao2019low} exploited a decoding manager to decide if the response is generated from three components, the context processor, the language model, and the knowledge processor. The context processor and language model can be trained without knowledge grounding labels, whereas the knowledge processor can be trained with few samples. \cite{wang2019improving} proposed different strategies for combining the context and the document, including sequential (like UniGDD), concatenate (like Fusion-in-Decoder), alternate (like \cite{li-etal-2019-incremental}), and interleave (some decoder attends to source, some to context). \cite{ghazvininejad2018knowledge} fuses the context embedding and the encodings of facts with a simple sum. Decoders with copy mechanism \cite{meng2020refnet,lin2020generating} build shortcuts between the knowledge text and the outputs.

Re2G \cite{glass-etal-2022-re2g} is closest to our approach. However, it is designed for generic knowledge augmented question answering or dialogue, ignoring the structural information widely available in documents, such as paragraphs, sequences, tables, etc. Some recent works have demonstrated the value of structural information, but they have only tried to use it in some modules of the DGDS pipeline \cite{fu2022doc2bot}, or in other tasks \cite{wan2021treejc}. But for \modelname{}, we integrate the structure into the entire processes of DGDS for the first time.

\begin{figure}
    \centering
    \includegraphics[width=0.98\textwidth]{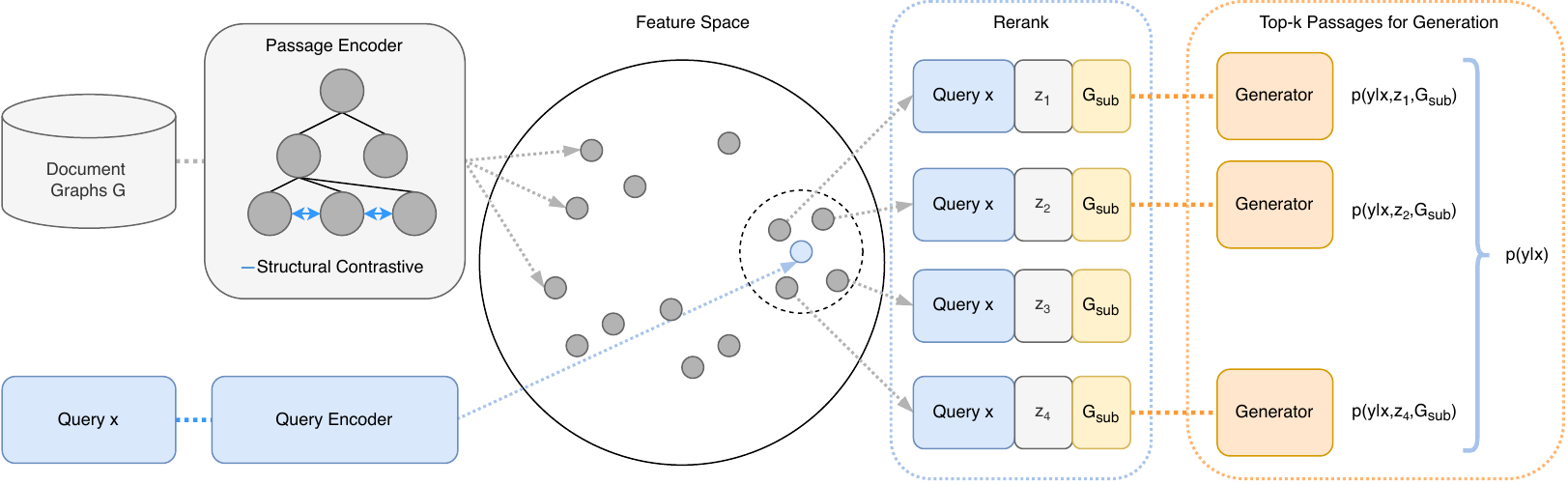}
    \caption{The architecture of {\modelname}. Like Re2G, SF-Re2G consists of a retrieval module, a rerank module, and a generation module. For retrieval, we design Structural Contrastive to enhance the negative sampling. For rerank and generation, we construct a subgraph for each passage to fuse more information. }
   \label{fig:sf-re2g-architecture}
\end{figure}

\section{Methodology}
The problem of DGDS can be described as follows. Assuming a collection of $D$ documents, we split each document into multiple passages and get $M$ total passages $\mathcal{Z}=\{z_1,...,z_M\}$. Given a conversational history $x=(u_1,a_1,...,u_{t-1},a_{t-1},u_t)$, where $u_i$ and $a_i$ are respectively the i-th user and agent utterances, our task is to generate the next system answer $a_t$ based on the passage collection. To fuse the structural information in the documents to DGDS, we transform each document into a document graph $G$, and the nodes of it will be regarded as the passages set $\mathcal{Z}$. For our approach, we will construct a subgraph $G_{\texttt{sub}}$ for each core passage, based on the whole document graph $G$.

The overall architecture of \modelname{} is shown in Figure \ref{fig:sf-re2g-architecture}. We follow the Retrieval-Rerank-Generation architecture. The retrieval module needs to efficiently find the top k passages relevant to a dialogue context. After that, a much more powerful rerank module will sort the retrieved passages more precisely. Finally, a generator will produce a response based on the top passages and their relevance scores. The detail of our method is described as follows.

\subsection{Retrieval}

The aim of the dense passage retrieval task is, given the user's query and all the passages from documents, to retrieve a set of passages that best matches the user's query. For each instance, a set of negative passages are required for the dense passage retrieval task. Through the comparison of positive and negative passages, the model can learn how to find the passages which best match the user's query. Therefore, the selection of negative passages will directly affect the performance of the model, and the selection of hard negatives is the most important among negative sampling.

The goal of the Retriever model is to efficiently find the top $k$ ($k<<M$) passages relevant to a dialogue context. We apply the Bi-encoder architecture for its efficiency. Specifically, a dialogue context $C$ and a passage candidate $p$ are first encoded by two separate BERT (base) encoders. The similarity scoring function is then defined as the dot product of the encoder outputs:
\begin{equation}
    \begin{split}
    P_{ret}(p|C)&\propto dot[BERT_c(C), BERT_p(p)]\\
    &=sim_{\eta_1}(C,p)
    \end{split}
\end{equation}

\paragraph{Training} Due to its high efficiency, BM25 is a common choice for sampling hard negative passages. However, considering the similarity of passages only based on the token by BM25 is far from enough for model learning. For the dense passage retrieval task, we utilize document structure to propose a more efficient hard negative sampling method that takes the semantic and logical similarity of the passages into account.

The document structure is a special kind of structure, which is usually a tree structure formed by artificially organizing documents and passages according to certain semantics or logic. The more semantically similar the two pieces of knowledge, the closer they are in this document structure tree space, which helps us find  hard negative sample easily on semantics and logic aspect.

To learn a better embedding function, we optimize the loss function as the negative log-likelihood of the positive passage:
\begin{gather*}
    \mathcal{L}(q_i, p_i^+, \{p_{i,j}^-\}_{j=1}^{m-1}, p_{i,m}^{h-})
\end{gather*}
{\small\begin{equation}
        = - log \frac{e^{sim(q, p_i^+)}}{e^{sim(q, p_i^+)} +\sum_{j=1}^{m-1} e^{sim(q, p_{i,j}^-)} + e^{sim(q, p_{i,m}^{h-})} }  \\
\end{equation}
}
where $q_i$ includes the current turn of user query and the dialog history between the user and the system, $p_i^+$ indicates one relevant (positive) passage, $\{p_{i,j}^-\}_{j=1}^{m-1}$ indicates in-batch negatives, $p_{i,m}^{h-}$ indicates hard negative, where $p_{i,m}^{h-}$ is sampled from the neighborhood of $p_i^+$ according to the document structure.

\begin{figure}
    \centering
    \includegraphics[width=0.98\textwidth]{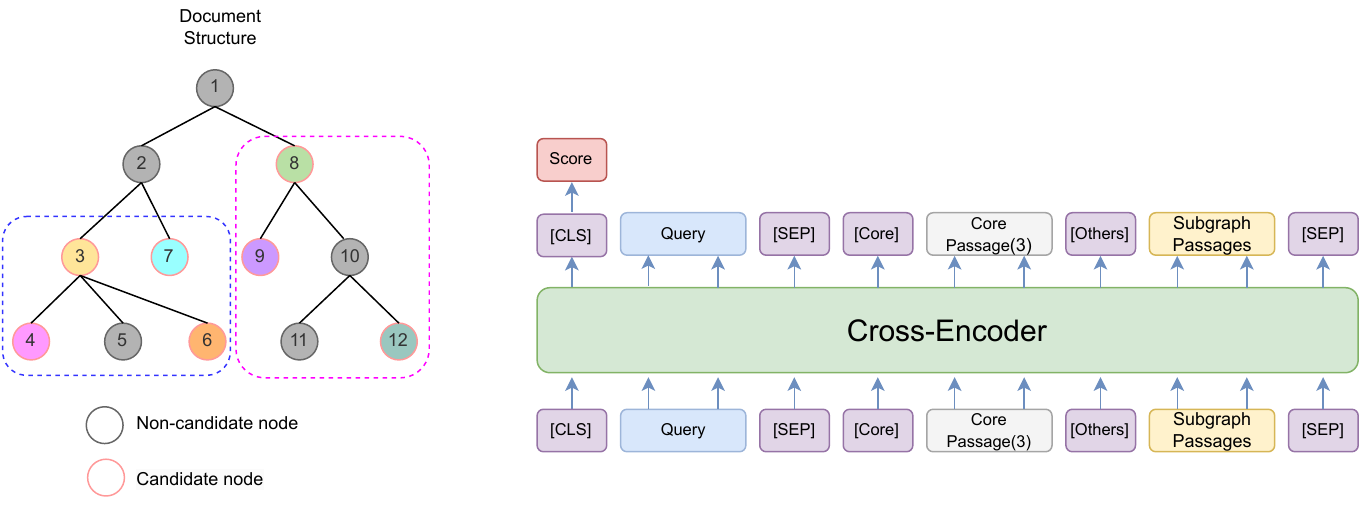}
    \caption{The architecture of the Reranker.}
    \label{fig:rerank-architecture}
\end{figure}
\subsection{Reranker}
\label{sec:rerank}
As a bridge between the retrieval module and generation module, on the one hand, a Reranker needs to rescore the candidates from the Retrieval through the cross-encoder structure, which is more time-consuming but more accurate than the bi-encoder structure to obtain a more effective score. On the other hand, the rerank rescores the candidates and directly affects the generation module.  

Based on the candidates from the Retrieval, the subgraph constructed according to the document structure, on the one hand, model the co-occurrence between different groundings. In the dataset, there are multiple grounding passages for one instance, like Doc2Bot\cite{fu2022doc2bot}. If the passage in the same group that has a high co-occurrence probability with the current passage matches the current question, the probability of the current passage should be improved. On the other hand, if the passage in the same group that does not have the possibility or has a low probability of co-occurrence matches the current question, the probability of the current passage being matched should be reduced.

Instead of inputting origin candidate passages \cite{DBLP:journals/corr/abs-1901-04085,10.1145/3477495.3531997} from the Retrieval, we first group all candidate passages from the Retrieval into different subgraphs according to some rules: 
\begin{enumerate}
    \item The passages in different documents will not appear in the same subgraph.
    \item The passages of the same document are grouped into the subgraph in a depth-first way to ensure that the distance of these passages in the hyperbolic space formed by the document structure is close enough.
    \item Determine whether to create a new subgraph according to the limited length of the input document and the length of the current subgraph. 

\end{enumerate}

Candidates from the retrieval are grouped into subgraphs according to the document structure and max input length. For scoring the candidate passages, we combine the origin passage with passages in the same subgraph to replace the origin passage. As is shown in Figure \ref{fig:rerank-architecture}, candidate passages nodes 3, 4, 6, 7, 8, 9, and 12 on this document form two clusters, where nodes 3, 4, 6 and 7 are in the same group and nodes 8, 9, and 12 are in the other group. The model input $\mathcal{X}$ for a passage $p_x$ of $l$ other passages in same subgraph becomes:

        $\mathcal{X}$ = [cls] query [sep] [core] $p_x$ [others] $p_1$ [sep] ...  $p_l$ [sep]

where `dialog' includes the current turn of user query and the dialog history between the user and the system. `[core]' is a special token indicating the start of the core passage. `[others]' is a special token indicating the start of other passages which is under the same document graph with the core passage.

A linear layer and softmax layer are then used on top of the [CLS] output of the cross-encoder model to obtain the similarity.

\paragraph{Training} Compared to the Retrieval, the Reranker is expected to rescore the candidates provided by the Retrieval and get a more accurate result through CrossEncoder architecture. The loss function for optimizing the Reranker is the same as the Retrieval: the negative log-likelihood of the positive passage.
\

\subsection{Generation}
For response generation, we consider the role of structural information in two ways. First, the structurally enhanced rerank scores mentioned in section \ref{sec:rerank} can reflect the difference in importance of the retrieved grounding passages. Second, similar to rerank, the information from neighbor nodes can help the generator better understand the content.

Based on these two considerations, we design a structure-fused weighting method for generation similar to \cite{patrick2020rag,glass-etal-2022-re2g,zhang2022retgen}. 
Differ from earlier interactive methods \cite{gao-etal-2022-unigdd} or later interactive methods \cite{lakhotia-etal-2021-fid}, the weighting method can directly leverage a grounded generation model trained on a single passage as input, retaining the knowledge in the pre-trained model as much as possible. It also yields flexibility in the number of top passages for the generation. Assuming the output of the generation model at step $i$ is $y_i$ and model parameters are $\theta$, we calculate the probability of the output token at the current step by:
{
\begin{equation}
    p(y_i|x) = \sum_{z\in top_k} p_\eta(z|x,G_{\texttt{sub}})p_\theta(y_i|x,z,G_{\texttt{sub}},y_{<i})
\end{equation}
}
Here, $p_\eta(\cdot|\cdot)$ is the output of the Rerank model, $\eta$ are the parameters it, and $G_{\texttt{sub}}$ is the subgraph around the retrieved passage $z$. In this way, we use structurally enhanced rerank scores to assign weights to each input passage, and our generative model can ground on not only the retrieved passage but also its context. In practice, we reuse the sub-graph $G_{\texttt{sub}}$ constructed in the rerank stage, making our method more convenient and efficient. We use a pre-trained BART \cite{lewis2020bart} model as the backbone network of the generator. The generator takes the concatenation of dialog history $x$, retrieved passage $z$, and the sequential sub-graph $G_{\texttt{sub}}$ as inputs. Moreover, the token sequence with the highest probability will be the output.

\paragraph{Training} Thanks to the feature of weight methods, we can easily implement the joint training of the reranker and generator without additional fine-tuning for the pre-trained BART model. End-to-end training can use the generated results to correct the weights given by the reranker, resulting in a better performance in the generation. Due to the limitation of model input sequence length, we will truncate the conversation history and keep only the most recent conversations. Between grounding passage $z$ and subgraph $G_{\texttt{sub}}$, we will prioritize keeping $z$ in the input.

\section{Experiments}
The implementation of our method is in PyTorch and the pre-trained models we used are from HuggingFace Transformers\footnote{\href{https://huggingface.co}{https://huggingface.co}}. The experimental details, including the hyperparameters and pre-trained models, can refer to the  S3 of the supplementary materials.

\subsection{Dataset}
We conduct experiments on MultiDoc2Dial \cite{feng2021multidoc2dial} and Doc2Bot \cite{fu2022doc2bot} datasets. One of them is in English, and another is in Chinese, which can verify the cross-linguistic ability of our model. The statistics of them are shown in Table \ref{tab:statistic}.

\begin{table}
    \centering
    \caption{Statistics of MultiDoc2Dial and Doc2Bot training set. Resp. Num. means the number of instances, Resp. Len. is the average length of system response.}
    {
    \small
    \begin{tabular}{c|ccc}
        \toprule
        Dataset & Domains & Resp. Num. & Resp. Len. \\
        \midrule 
        \multirow{4}*{MultiDoc2Dial} & ssa & 4,993 & 18.80 \\
        & va & 6,238 & 20.81 \\
        & dmv & 6,135 & 21.39 \\
        & student & 4,085 & 20.37 \\
        \midrule
        \multirow{4}*{Doc2Bot} & health & 1,241 & 39.60\\
        & tech. & 1,181 & 64.74\\
        & insurance & 1,471 & 67.71\\
        & wikihow & 1,856 & 61.47\\
        \bottomrule
    \end{tabular}
    }
    \label{tab:statistic}
\end{table}

\begin{description}
    \item[MultiDoc2Dial] provides a training set of 3,474 dialogs corresponding to 48,002 utterances and 661 dialogs corresponding to 9,195 utterances for evaluation. The documents are written in English and collected from 4 domains. 
    \item[Doc2Bot] provides a training set of 800 dialogs corresponding to 11,485 utterances and 200 dialogs corresponding to 2,941 utterances for evaluation until now. The documents of it are written in Chinese and collected from 4 domains. 
\end{description}



\subsection{Retrieval and Reranking}
\begin{table}
    \centering
    \caption{Ablation study result of Retrieval and Rerank on MultiDoc2Dial and Doc2Bot.}
    {
        \begin{tabular}{l|cccc|cccc}
        \toprule
        \multirow{2}*{Method} & \multicolumn{4}{c|}{MultiDoc2Dial} & \multicolumn{4}{c}{Doc2Bot}\\ \cmidrule{2-5} \cmidrule{6-9}
         & \textbf{R@1} & \textbf{R@5} & \textbf{R@10}& \textbf{R@100} &  \textbf{R@1} & \textbf{R@5} & \textbf{R@10}& \textbf{R@100}\\ 
        \midrule
        BM25 &  0.136 & 0.312 & 0.406 & 0.727 & 0.067 & 0.178 & 0.250 & 0.581\\
        DPR  & 0.402 & 0.661 & 0.745 & 0.925 & 0.468 & 0.713 & 0.798 & \textbf{0.950} \\
        DPR$_{\texttt{struct}}$ (ours) & \textbf{0.424} & \textbf{0.682} & \textbf{0.771} & \textbf{0.938} & \textbf{0.486} & \textbf{0.723} & \textbf{0.807} & \textbf{0.950} \\
        \midrule
         & \textbf{R@1} & \textbf{R@5} & \textbf{R@10}& \textbf{Rprec} &  \textbf{R@1} & \textbf{R@5} & \textbf{R@10}& \textbf{Rprec}\\ 
         \midrule
        CrossEncoder$_{\texttt{struct}}$ (ours) & \textbf{0.550} & 0.783 & 0.832 & \textbf{0.553} & \textbf{0.559} & \textbf{0.757} & 0.820 & \textbf{0.624}\\
         CrossEncoder & 0.547 & \textbf{0.785} & \textbf{0.841} & 0.550 & 0.550 & 0.741 & \textbf{0.823} & 0.606 \\
        \bottomrule
    \end{tabular}
    }
    \label{tab:retrieve}
\end{table}

\paragraph{Metrics}  We use Recall@1, Recall@5, Recall@10 and Recall@100 to evaluate the retrieval result. Because the input of the Reranker is from top@100 from the retrieval, Recall@100 for each rerank model is same. We use Recall@1, Recall@5, Recall@10 and Rprec to evaluate the rerank result.

\paragraph{Baselines} For the retrieval part, we apply a common but efficient method, DPR, to ensure the comparability of the final experiment. Inspired by TreeJC \cite{wan2021treejc}, we modify the baseline by adding top-down aggregation to the passage to enhance the representation of passages using upper-level information such as paragraph headings and document names. We conduct an ablation study to validate the effectiveness of our proposed structure-based negative sampling strategy.  

For the rerank part, we apply a finetune CrossEncoder pretraining model, RoBERTa. Inspired by TreeJC \cite{wan2021treejc}, we also add the top-down aggregation to the passage for the input of RoBERTa to build a strong baseline model. 

\paragraph{Discussion} As we can see from Table \ref{tab:retrieve}, on both the MultiDoc2Dial dataset and Doc2Bot dataset, we can see the performance will also drop significantly: Averagely, on MultDoc2Dial, there is about 2\% drop; on doc2bot, there is about 1\% drop. We conduct an ablation study to validate the effectiveness of adding the structure information on Reranker. As we can see from Table \ref{tab:retrieve}, on the Doc2Bot dataset, by adding the structure information on Reranker, there is about a 1\% improvement on every metric on average, but on the MultiDoc2Dial dataset with less structural information, no obvious change appears on evaluation metrics.

\subsection{Response Generation} 
\begin{table}
    \centering
    \caption{Generation results of compared methods on MultiDoc2Dial and Doc2Bot. We replaced the generator in Re2G with Fusion-in-Decoder to form Re2FiD. \texttt{w/o subgraph} means removing the subgraph in the generation stage while keeping the retrieval and rerank unchanged.}
    {
        \begin{tabular}{l|ccc|ccc}
        \toprule
        \multirow{2}*{Method} & \multicolumn{3}{c|}{MultiDoc2Dial} & \multicolumn{3}{c}{Doc2Bot}\\ \cmidrule{2-4} \cmidrule{5-7} 
         & \textbf{F1} & \textbf{S-BLEU} & \textbf{ROUGE}& \textbf{F1} & \textbf{S-BLEU} & \textbf{ROUGE}\\ 
        \midrule
        RAG \cite{patrick2020rag} & 35.92 & 17.54 & 32.41  & 49.39 & 30.21 & 49.64 \\
        Re2G \cite{glass-etal-2022-re2g} & 42.30 & 22.78 & 38.08 & 54.53 & 36.17 & 54.78\\
        Re2FiD* \cite{lakhotia-etal-2021-fid} & 42.22 & 23.35 & 38.39 & \textbf{57.09} & \textbf{38.31} & 55.10\\
        \midrule
        \modelname{} (ours)  & 42.13 & 23.06 & 37.95 & \textbf{57.04} & 37.49 & \textbf{57.38} \\
        \modelname{}$_{\texttt{w/o subgraph}}$ (ours) & \textbf{43.65} & \textbf{24.12} & \textbf{39.62} & 55.85 & 36.05 & 55.97 \\
        \bottomrule
    \end{tabular}
    }
    \label{tab:generation}
\end{table}

\paragraph{Metrics} We use token-level F1 score, SacreBLEU \cite{post-2018-call}, and ROUGE \cite{lin2004rouge} to evaluate the response. The code of them is similar to MultiDoc2Dial\footnote{\href{https://github.com/IBM/multidoc2dial}{https://github.com/IBM/multidoc2dial}}. For the Chinese dataset Doc2Bot, we regard every character as a token.

\paragraph{Baselines} We use the recent SOTA methods, including RAG \cite{patrick2020rag}, Re2G \cite{glass-etal-2022-re2g} and Fusion-in-Decoder \cite{lakhotia-etal-2021-fid}, as baselines. For more details of these methods, please refer to S2 in the supplementary materials.

\paragraph{Discussion} Table \ref{tab:generation} shows the generation results. As we can see, our \modelname{} can obtain SOTA performance under multiple metrics for both datasets. It is worth noting that our generator uses only half of the parameters of Re2FiD and still achieves better performance. The interesting thing is that the generation performance drops about 1 point for all evaluation metrics in Doc2Bot dataset after removing the subgraph from the input sequence, but in MultiDoc2Dial the performance improves. We conjecture that on the dataset with insignificant structural information like MultiDoc2Dial, subgraphs do not lead to better results for the generation. Instead, using the original passage with our structure enhanced rerank yields higher performance.
 
 \section{Conclusion}
This paper presented {\modelname}, a novel approach that uses the structural information in the document source to facilitate DGDS. Unlike prior works, which focus solely
on the textual content of the document knowledge, {\modelname} systematically investigates the role of structural information on DGDS. We design different methods to fuse the structural information into the different stages of DGDS pipeline. Our experimental results showed that: 1) the hard negative samples collected from the document structure are more difficult to distinguish and meaningful to distinguish than those collected based on BM25, which can bring about 1 \% or more improvement on both the MultiDoc2Dial dataset and Doc2Bot dataset 2) For the candidate of retrieval, the subgraph is built based on the document structure, which can increase the matching and comparison information so that the rerank can be improved by about 1\%. 3) the subgraph information could be helpful if the original documents have some structures. We hope that our approach and such observations will raise awareness of the importance of structural information and be helpful for future research in this direction.

\bibliography{anthology,custom}
\bibliographystyle{splncs04}

\end{document}


\title{Supplementary Material:\\ Structures Facilitate Retrieve, Rerank, and Generate}
\author{Anonymous Author(s)}
\authorrunning{Anonymous Author(s)}
\institute{Anonymous Institution(s) \\
\email{anonymous@example.com}}
\maketitle

\section{Supplementary Related Works}

This part provides the detailed descriptions of knowledge selection and multi-turn dialog representation methods that are only briefly mentioned in the main paper. The content is organized following the same subsection structure.

\subsection{Extended Knowledge Selection Methods}

Beyond the methods discussed in the main paper, further approaches have been proposed to improve knowledge selection. RocketQA \cite{yingqi2021rocketqa} and RocketQA2 \cite{ren2021rocketqav2} improve the negative sampling stage of DPR and apply end-to-end training between the retrieval module and rerank module. ColBert \cite{khattab2020colbert} performs later interaction on the encoded features, but it can be a trade-off between performance and efficiency. Specifically, some methods \cite{chen-etal-2020-bridging,meng2020dukenet,kim2019sequential,lian2019learning} try to improve knowledge selection by reducing the gap between prior and posterior distributions. KnowledgeGPT \cite{zhao-etal-2020-knowledge-grounded} exploits a sequential knowledge selection module. Noticeably, the knowledge selection module is first trained using pseudo-labels, which use the response to select knowledge, then fine-tuned on dialogs with knowledge annotated using curriculum training. By doing so, the model can benefit from unlabeled dialogues, reducing the annotation effort. DiffKS \cite{zheng-etal-2020-difference} tries to improve the diversity of different turns while maintaining consistency by simultaneously modeling the difference and similarity for knowledge selection.

\subsection{Extended Multi-turn Dialog Representation Methods}

Additional multi-turn dialog matching methods are detailed here. MSN \cite{yuan-etal-2019-multi} performs 2-phase matching: the first one filters multi-hop utterance selection, and the second phase matches filtered utterances to the response. IMN \cite{gu2019interactive} reduces noise in context \cite{ma-etal-2020-compare} using a compare-and-aggregate strategy. The drawback of considering such multi-turn matching is that the retrieval model is inefficient.




\section{Details of the Sota method}

\begin{description}
    \item[RAG] follows the Retrieval-and-Generation architecture, where we use BERT-base for retrieval and BART-large for response generation. RAG introduces relevance weights into text generation at the token level, which is similar to our method. In this way, the higher relevant passage has more influence on the response generation.

    \item[Re2G] consists of a retrieval module, a rerank module, and a generation module. The retrieval module uses a dual-encoder architecture \cite{karpukhin2020dpr}, using two encoders to encode the user query and the passages in the document source. FAISS \cite{johnson2019faiss} is used to index the passages, retrieving the top-k related passages based on the dot similarity. After that, the cross-encoder rerank module will give each retrieved document a more accurate relevance score. Finally, the generator will produce the response based on the top passages from the rerank module and their rerank score. Here, we use the same training steps as in \cite{glass-etal-2022-re2g} for the experiment. We only changed some hyper-parameters to fit our datasets. Specifically, we use BERT-base for retrieval, RoBERTa-large for reranking, and BART-large for generation. The generation is based on relevance weights similar to RAG. The only difference is that the relevance score is from rerank, not retrieval.

    \item[Re2FiD] is a modification of Re2G, where the retriever and the reranker are the same as Re2G. Specifically, we use a BERT-base for retrieval, a RoBERTa-large for reranking, and a T5-large \cite{colin2020t5} for generation. Unlike RAG and Re2G, the generator of it follows the Fusion-in-Decoder (FiD) approach. FiD encodes each passage independently. Then the decoder takes the concatenation of the encoded passages as input.
    
    \item[{\modelname}] also follows the Retrieval-Rerank-Generation architecture. We use a pre-trained BART-large as the generator. The detail of our method has been described in the above sections. To verify the importance of the subgraph for generation, we also conduct ablation experiments by removing $G_{\texttt{sub}}$ in the input sequence while keeping the retrieval and rerank unchanged.
\end{description}





\section{Experimental Details}
\label{sec:appendix-experiments}
\begin{table}
    \centering
    \begin{tabular}{c|c|ccc}
        \toprule \textbf{Language} & \textbf{Model} & \textbf{Variant} & \textbf{Parameters} & \textbf{HuggingFace Pre-trained Model} \\
         \midrule
         \multirow{5}*{English}
         & DPR$_\texttt{qry}$ & base & 109M & \small\tt{facebook/dpr-question\_encoder-multiset-base}  \\
         & DPR$_\texttt{ctx}$ & base & 109M & \small\tt{facebook/dpr-ctx\_encoder-multiset-base}  \\
         & RoBERTa & large & 355M & \small\tt{roberta-large}\\
         & BART & large & 406M & \small\tt{facebook/bart-large}\\
         & T5 & large & 738M & \small\tt{t5-large}\\
         \midrule
         \multirow{5}*{Chinese}
         & RoBERTa$_\texttt{qry}$ & base & 102M & \small\tt{hfl/chinese-roberta-wwm-ext} \\
         & RoBERTa$_\texttt{ctx}$ & base & 102M & \small\tt{hfl/chinese-roberta-wwm-ext}\\
         & RoBERTa & large & 326M & \small\tt{hfl/chinese-roberta-wwm-ext-large} \\
         & BART & large & 407M & \small\tt{fnlp/bart-large-chinese}\\
         & T5 & large & 751M & \small\tt{IDEA-CCNL/Randeng-T5-784M}\\
         \toprule
    \end{tabular}
    \caption{The detail of the pre-trained model for all baselines. Since there is no pre-trained DPR in Chinese, we use RoBERTa as a substitute.}
    \label{tab:model}
\end{table}
 The information of the pre-trained model we used is shown in Table \ref{tab:model}. All experiments are performed on one Tesla A100 with 80GB memory. The hyper-parameters for each training stage are shown in the follows:
 \subsection{Hyperparameters for Retrieval}
 For the DPR, we use a pre-trained RoBERTa as the backbone network. We train the model for 20 epochs with a batch size of 128. Gradient checkpoint is used to break the GPU memory limitation. The learning rate is 3e-5, and linear scheduling with warming up is applied. We use one BM25 negative passage per query in addition to in-batch negatives. Furthermore, the  FAISS\footnote{\href{https://github.com/facebookresearch/faiss}{https://github.com/facebookresearch/faiss}} is used to speed up the vector search. 
 \subsection{Hyperparameters for Rerank}
 We use a pre-trained RoBERTa as the backbone network for reranking. We use the top-20 retrieved samples, except for the gold ones, as negative samples for training. We train it for five epochs with a learning rate of 8e-6 on MultiDoc2Dial and a learning rate of 3e-5 on Doc2Bot, which is linear scheduled with a warmup. For inference, we use the top-100 retrieved samples as candidates.
 
 \subsection{Hyperparameters for Generation}
 We directly apply the end-to-end training for our rerank module with a pre-trained BART from HuggingFace. We will truncate the dialog history to 195 tokens due to the model limitation. We train the joint model for 10 epochs with a batch size of 8. And the learning rate is 2e-5, with linear scheduling of 500 warmup steps. For inference, we set the beam search size as 3, and the max generate length as 512. The variants of it use the same hyperparameters since the only difference is the input sequence.

\bibliography{anthology,custom}
\bibliographystyle{splncs04}